%% file: acl_latex.tex
\documentclass[11pt]{article}

\usepackage[preprint]{acl}

\usepackage{times}
\usepackage{latexsym}

\usepackage[T1]{fontenc}

\usepackage[utf8]{inputenc}

\usepackage{microtype}

\usepackage{inconsolata}

\usepackage{graphicx}

\usepackage{float}
\usepackage{fancyvrb}
\usepackage{graphicx}
\usepackage{boldline} 
\usepackage{booktabs}
\usepackage{multirow}
\usepackage{sidecap}
\usepackage{subcaption}
\usepackage{amsmath}
\usepackage{amsfonts}
\usepackage{xcolor}
\usepackage{bm}
\usepackage{hyperref}
\usepackage{rotating}
\usepackage{adjustbox}
\usepackage{algorithm}
\usepackage{algpseudocode} 
\usepackage{array}
\definecolor{green}{rgb}{0.0, 0.5, 0.0}
\definecolor{red}{rgb}{0.82, 0.1, 0.26}
\definecolor{codegray}{gray}{0.9}
\newcommand{\code}[1]{\texttt{#1}}

\usepackage{listings}
\definecolor{codegreen}{rgb}{0,0.6,0}
\definecolor{codegray}{rgb}{0.5,0.5,0.5}
\definecolor{codepurple}{rgb}{0.58,0,0.82}
\definecolor{backcolour}{rgb}{0.95,0.95,0.92}
\lstdefinestyle{mystyle}{
    backgroundcolor=\color{backcolour},   
    commentstyle=\color{codegreen},
    keywordstyle=\color{magenta},
    numberstyle=\tiny\color{codegray},
    stringstyle=\color{codepurple},
    basicstyle=\ttfamily\footnotesize,
    breakatwhitespace=false,         
    breaklines=true,                 
    captionpos=b,                    
    keepspaces=true,                 
    numbers=left,                    
    numbersep=5pt,                  
    showspaces=false,                
    showstringspaces=false,
    showtabs=false,                  
    tabsize=2
}
\usepackage{pifont}
\newcommand{\cmark}{\ding{51}}%
\newcommand{\xmark}{\ding{55}}%

\newlength{\Oldarrayrulewidth}

\definecolor{darkpurple}{RGB}{75, 0, 130}
\definecolor{darkyellow}{RGB}{230, 184, 0}

\usepackage[most]{tcolorbox}
\tcbuselibrary{breakable}   

\newtcolorbox{myshadowbox}{
    enhanced,
    colback=white, 
    colframe=black, 
    shadow={1mm}{-1mm}{0mm}{black!50!white}, 
    boxrule=0.5pt 
}
\usepackage{colortbl} 
\usepackage{tabularx} 

\newcounter{prop}

\newif\ifcoloredtext
\coloredtextfalse 

\usepackage{amsthm}
\newtheorem*{assumption*}{Assumption}

\usepackage{tikz}
\usetikzlibrary{arrows.meta, positioning, shapes.multipart, calc}

\title{LiTS: A Modular Framework for LLM Tree Search}

\author{Xinzhe Li \\
  RMIT University \\
  \texttt{xinzhe.li@rmit.edu.au} \\\And
  Yaguang Tao \\
  RMIT University \\
  \texttt{yaguang.tao@rmit.edu.au} \\}

\begin{document}
\maketitle

\input{sections/abstract}
\input{sections/1.intro}

\input{sections/3.1.lits_core}
\input{sections/3.2.lits_tool}

\input{sections/4.eval}

\input{sections/2.related}

\input{sections/5.conclusion}

\bibliography{custom}
\clearpage
\appendix

\input{sections/6.1.app-save-dir}

\input{sections/6.2.app-exp-additional}

\end{document}

%% file: sections/abstract.tex
\begin{abstract}
LiTS is a modular Python framework for LLM reasoning via tree search. It decomposes tree search into three reusable components—Policy, Transition, and RewardModel—that plug into algorithms like MCTS and BFS. A decorator-based registry enables domain experts to extend to new domains by registering components, and algorithmic researchers to implement custom search algorithms. We demonstrate composability on MATH500 (language reasoning), Crosswords (environment planning), and MapEval (tool use), showing that components and algorithms are orthogonal: components are reusable across algorithms within each task type, and algorithms work across all components and domains.
We also report a mode-collapse finding: in infinite action spaces, LLM policy diversity—not reward quality—is the bottleneck for effective tree search. A demonstration video is available at \url{https://youtu.be/nRGX43YrR3I}.
The package is released under the Apache 2.0 license at \url{https://github.com/xinzhel/lits-llm}, including installation instructions and runnable examples that enable users to reproduce the demonstrated workflows.

\end{abstract}

%% file: sections/1.intro.tex
\input{tabs/lits_architechture}

\section{Introduction}
Tree search methods for LLM inference—such as 
Tree-of-Thoughts \citep{yao2023tree}, 
RAP \citep{hao-etal-2023-reasoning}, and 
ReST-MCTS \citep{zhang2024restmcts}—have demonstrated 
strong performance on complex reasoning tasks.
However, existing implementations are largely task-specific, 
requiring substantial reimplementation effort 
when adapting to new domains or comparing across methods.

We present \textbf{LiTS} (Language Inference via Tree Search), 
a modular Python framework designed to let users \textit{modify what they need and reuse the rest}.
LiTS targets two user groups with complementary workflows:
\begin{itemize}
\item \textbf{AI/NLP researchers} devise novel search algorithms for direct evaluation on downstream tasks, 
or design new reasoning structures (e.g., sub-question decomposition~\citep{hao-etal-2023-reasoning})—
changes to step structure cascade only to dependent abstractions (detailed in §\ref{sec:lits}), reusing search algorithms.

\item \textbf{Domain experts} inject domain-specific logic (prompts, tools, environment dynamics) without touching search internals.
\end{itemize}

LiTS achieves this by isolating inference algorithms from domain-specific implementations, enabling bidirectional reusability (Table~\ref{tab:reusability}):
domain experts can compare different methods (MCTS, BFS, CoT) with identical domain components for fair algorithmic comparison;
researchers can evaluate the same method across domains by swapping only domain-specific components.

Correspondingly, LiTS also addresses three key challenges 
in developing LLM reasoning agents.
\begin{enumerate}
\item \textbf{Reusability}: 
General data structures (State, Action, Step) decouple inference algorithms from domain implementations, enabling bidirectional sharing (Table~\ref{tab:reusability}). Domain-specific logic is further decomposed into three fine-grained components---Policy, Transition, and RewardModel~\citep{li2024survey}. As a result, domain experts reuse search algorithms across domains by swapping only domain-specific components; researchers evaluate the same algorithm across domains with identical domain logic. The package design would be further demonstrated in \S\ref{sec:lits}.

\item \textbf{Extensibility}: Components and search modules can be extended via external import and decorator-based registration, without modifying the core package.
Table~\ref{tab:extension} exemplifies extension patterns which would further demonstrated in \S\ref{sec:demo}.


\item \textbf{Observability}: 
Built-in \code{InferenceLogger} tracks token usage and latency at component, instance, and search-phase granularities, plus incremental checkpointing for post-training.
\end{enumerate}

The accompanying video demonstrates both modular configuration workflows and a brief live execution illustrating real tree-search behavior.

\begin{table}[h]
\centering
\small
\begin{tabular}{p{1.36cm}p{2.6cm}p{2.55cm}}
\toprule
\textbf{User Group} & \textbf{Reusability}\newline \textbf{Direction} & \textbf{Benefit} \\
\midrule
Domain experts & Same components 
\newline $\rightarrow$ different methods 
& Fair algorithmic comparison 
\\ \addlinespace

AI/NLP \newline researchers 
& Same method 
\newline $\rightarrow$ different domains 
& Generalization testing 
\\\addlinespace

AI/NLP \newline researchers 
& New formulation \newline $\rightarrow$ existing search 
& Novel reasoning paradigms \\
\bottomrule
\end{tabular}
\caption{Bidirectional reusability enabled by LiTS's modular design.}
\label{tab:reusability}
\end{table}

\begin{table*}[ht!]
\centering
\small
\begin{tabular}{p{2.2cm}p{3.2cm}p{4.8cm}p{3.5cm}}
\toprule
\textbf{Task Type} & \textbf{Minimum Required Injection} & \textbf{Extension Mechanism} & \textbf{Framework Knowledge} \\
\midrule
Language-grounded & Prompt, Dataset & \code{@register\_user\_prompt}, \code{@register\_dataset} & None \\
Env-grounded & Transition class & \code{@register\_transition} (1 decorator) & Minimal\textsuperscript{$\ddagger$} \\
Tool-use & Tool definitions & \code{BaseTool} subclass & Minimal\textsuperscript{*} \\
\bottomrule
\end{tabular}
\caption{Domain extension patterns by task type. Component reuse varies: tool-use tasks share Policy and Transition across domains; env-grounded tasks share Policy and RewardModel; language-grounded tasks may require task-specific prompts. \textsuperscript{$\ddagger$}Requires understanding Transition interface (\code{goal\_check}, \code{generate\_actions}, \code{\_step}). \textsuperscript{*}\code{BaseTool} requires \code{name}, \code{description}, \code{args\_schema}, and \code{\_run()}—compatible with LangChain's Tool interface.}
\label{tab:extension}
\end{table*}

%% file: tabs/lits_architechture.tex
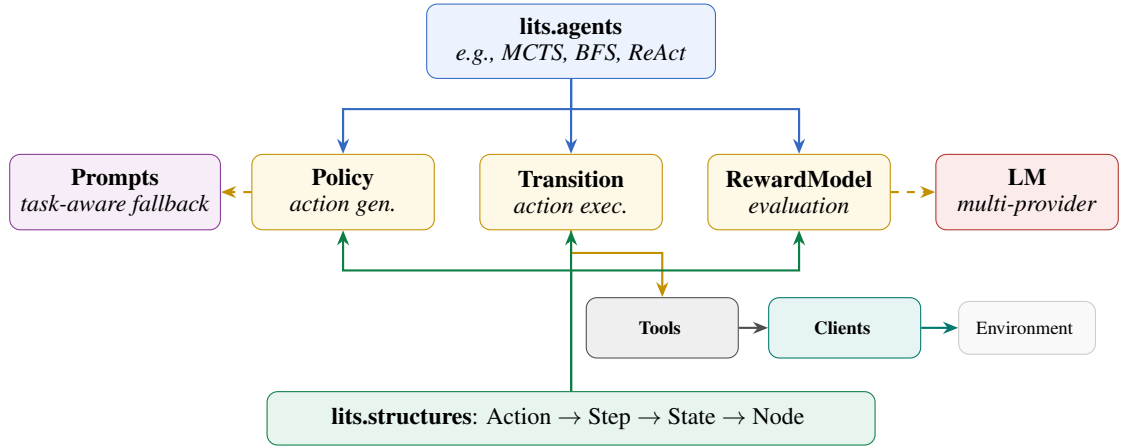
\begin{figure*}[t]
\centering
\definecolor{agentblue}{RGB}{66,133,244}
\definecolor{componentorange}{RGB}{244,180,0}
\definecolor{structgreen}{RGB}{15,157,88}
\definecolor{promptpurple}{RGB}{171,71,188}
\definecolor{toolgray}{RGB}{97,97,97}
\definecolor{clientteal}{RGB}{0,150,136}
\definecolor{lmred}{RGB}{219,68,55}

\tikzset{
    layerbox/.style={rounded corners, draw=#1!80!black, fill=#1!10, 
        minimum width=3.8cm, align=center, inner sep=5pt, font=\footnotesize},
    widebox/.style={rounded corners, draw=#1!80!black, fill=#1!10, 
        minimum width=8cm, align=center, inner sep=6pt, font=\footnotesize},
    smallbox/.style={rounded corners, draw=#1!80!black, fill=#1!10, 
        minimum width=2.4cm, minimum height=1cm, align=center, inner sep=4pt, font=\footnotesize},
    tinybox/.style={rounded corners, draw=#1!80!black, fill=#1!10, 
        minimum width=2cm, minimum height=0.8cm, align=center, inner sep=3pt, font=\scriptsize},
    arrow/.style={-{Stealth[length=2mm,width=1.6mm]}, thick, draw=#1!80!black},
    dashedarrow/.style={-{Stealth[length=2mm,width=1.6mm]}, thick, draw=#1!80!black, dashed}
}

\begin{tikzpicture}

\node[layerbox=agentblue] (agent) at (0, 0) {
    \textbf{lits.agents}\\
    \textit{e.g., MCTS, BFS, ReAct}
};

\node[smallbox=componentorange] (policy) at (-3, -2) {
    \textbf{Policy}\\
    \textit{action gen.}
};

\node[smallbox=componentorange] (transition) at (0, -2) {
    \textbf{Transition}\\
    \textit{action exec.}
};

\node[smallbox=componentorange] (reward) at (3, -2) {
    \textbf{RewardModel}\\
    \textit{evaluation}
};

\node[smallbox=promptpurple] (prompt) at (-6, -2) {
    \textbf{Prompts}\\
    \textit{task-aware fallback}
};

\node[smallbox=lmred] (lm) at (6, -2) {
    \textbf{LM}\\
    \textit{multi-provider}
};

\node[tinybox=toolgray] (tools) at (1.2, -3.8) {
    \textbf{Tools}
};

\node[tinybox=clientteal] (clients) at (3.6, -3.8) {
    \textbf{Clients}
};

\node[draw=gray!40, rounded corners, align=center, font=\scriptsize, 
    fill=gray!5, minimum width=1.8cm, minimum height=0.7cm] (env) at (6.0, -3.8) {
    Environment
};

\node[widebox=structgreen] (structure) at (0, -5) {
    \textbf{lits.structures}: Action $\rightarrow$ Step $\rightarrow$ State $\rightarrow$ Node
};


\draw[arrow=agentblue] (agent.south) -- ++(0,-0.4) -| (policy.north);
\draw[arrow=agentblue] (agent.south) -- (transition.north);
\draw[arrow=agentblue] (agent.south) -- ++(0,-0.4) -| (reward.north);

\draw[dashedarrow=componentorange] (policy.west) -- (prompt.east);

\draw[dashedarrow=componentorange] (reward.east) -- (lm.west);

\draw[arrow=componentorange] (transition.south) -- ++(0,-0.3) -| (tools.north);

\draw[arrow=toolgray] (tools.east) -- (clients.west);
\draw[arrow=clientteal] (clients.east) -- (env.west);

\draw[arrow=structgreen] (structure.north) -- ++(0,1.6) -| (policy.south);
\draw[arrow=structgreen] (structure.north) -- ++(0,0.8) -- (transition.south);
\draw[arrow=structgreen] (structure.north) -- ++(0,1.6) -| (reward.south);

\end{tikzpicture}

\caption{%
LiTS architecture. 
\textcolor{agentblue}{Agents} orchestrate \textcolor{componentorange}{Components} (Policy, Transition, RewardModel). 
Components use \textcolor{promptpurple}{Prompts} and \textcolor{lmred}{LM} interfaces.
Transition can invoke \textcolor{toolgray}{Tools} via \textcolor{clientteal}{Clients} for tool-use tasks.
All layers communicate through shared \textcolor{structgreen}{Data Structures}, decoupling search algorithms from domain logic.
}
\label{fig:lits-architecture}
\end{figure*}

%% file: sections/3.1.lits_core.tex
\section{LiTS: A Unified Grammar}
\label{sec:lits}
LiTS abstracts three task types—
\textbf{Environment Grounded} (e.g., BlocksWorld), 
\textbf{Language Grounded} (e.g., Math QA), 
and \textbf{Tool Use} (e.g., MapEval)—to 
unify diverse LLM reasoning paradigms.

Figure~\ref{fig:lits-architecture} presents the architecture. 
Each module is a separately deployable unit 
with dependencies on adjacent modules.

\subsection{lits.structures: General Data Structures for Decoupling}
The framework relies on a hierarchical type system to separate inference algorithms from domain-specific implementations: 
Action $\rightarrow$ Step $\rightarrow$ State $\rightarrow$ Node. 
Each level defines:
(1) a \textbf{task-agnostic interface} that enables \code{lits.agents} to operate without knowing underlying implementations, and
\begin{itemize}
\item \textbf{Action} (general interface): 
The atomic unit generated by Policy.

\item \textbf{Step} (general interface): 
Encapsulates an action with its execution results (if any).

\item \textbf{State} (general interface): 
Accumulation of steps over time, with unified rendering methods.

\item \textbf{Node} (general interface): 
Search-oriented wrapper managing parent pointers, children, 
rewards, and visit counts for search algorithms.
\end{itemize}
(2) \textit{polymorphic subclasses} for message passing between task-specific components (see Appendix Table~\ref{tab:structure_abstraction} for concrete subclasses across task types).

This design decouples search algorithms from task-specific logic. 
Researchers implement new reasoning domains by defining 
task-specific Action, Step, and State subclasses, 
while reusing the same search pipelines (BFS, MCTS, etc.).


\subsection{Modular LLM Components (lits.components) }
LiTS decomposes LLM-based reasoning into three modular components, defined by a \textbf{general interface} with 
task-specific subclasses:
\textbf{Policy} generates candidate actions from states .
\textbf{RewardModel} evaluates action quality for search guidance .
\textbf{Transition} executes actions and produces new states.

\paragraph{Provided Implementations.}
Most components are designed for a general task type 
(as are the prompts described below).
They declare a \code{TASK\_TYPE} class constant 
(e.g., \code{'language\_grounded'}, \code{'tool\_use'}, 
\code{'env\_grounded'}).
For example, both math reasoning and commonsense reasoning 
can use the same components, along with default prompts.

However, component design is not limited to the predefined types. 
For task instances requiring specialized behavior, 
developers can create task-instance-specific components 
with \code{TASK\_TYPE = None}.
Our package provides \code{BlocksWorldTransition} as an example, 
which implements BlocksWorld-specific state parsing and goal checking.
Subclass implementations focus solely on task-specific logic; infrastructure concerns such as inference logging, prompt registry lookup, and search-phase context are handled by base classes.

\paragraph{Decorator-Based Component Extension.}
LiTS uses a \code{ComponentRegistry} with decorator-based registration
so that users extend the framework without modifying core code.
For environment-grounded tasks, users register a \code{Transition}
subclass (implementing \code{\_step}, \code{goal\_check}, and
\code{generate\_actions}) via \code{@register\_transition("name")}
alongside a dataset loader via \code{@register\_dataset("name")};
by convention both share the same key, so the factory resolves
the transition from the dataset name
(e.g., \code{--dataset crosswords}; see \S4.1).
For language-grounded tasks, a \textit{search framework}
(e.g., \code{--search\_framework rap}) bundles a matching
Policy, Transition, and RewardModel from the registry,
while users may also define custom Step subclasses
for novel reasoning formulations (see \S4.3).
For tool-use tasks, users define tools via \code{BaseTool}
subclasses; the generic \code{ToolUsePolicy} and
\code{ToolUseTransition} handle orchestration (see \S4.2).
In all cases, users can independently specify
\code{--policy}, \code{--transition}, and \code{--reward}
on the CLI to compose custom component configurations
without writing new code—effectively assembling
their own search formulations from registered building blocks
(see \S4.3 for a concrete example).
This differentiates LiTS from LLM Reasoners~\citep{hao2024llm}:
rather than requiring task-specific component implementations for each search method,
LiTS provides a plug-and-play extension mechanism
where domain experts focus solely on domain logic.

\subsection{lits.prompts: Task-Instance-Prioritized Fallback.}
LiTS uses a centralized \code{PromptRegistry} to manage prompts 
for all LLM-based components, 
enabling prompt reuse and customization.
Components use two prompt types: 
\code{task\_prompt\_spec} (system instructions) and 
\code{usr\_prompt\_spec} (user message templates).

The registry supports a fallback lookup priority: 
explicit parameter $\rightarrow$ \code{task\_name} $\rightarrow$ 
\code{TASK\_TYPE} $\rightarrow$ default.
This allows users to:
(1) reuse general task-type prompts for new task instances 
(e.g., a new math dataset can use \code{language\_grounded} prompts), or
(2) register task-instance-specific prompts when needed.
Task-instance-specific components should set \code{TASK\_TYPE = None}
to skip the \code{TASK\_TYPE} fallback, 
avoiding potential mismatch between prompts and parsing logic.

\subsection{Agents}
LiTS provides chain agents (\code{ReActChat}, \code{EnvChain}) for sequential reasoning 
and tree search agents (\code{MCTS}, \code{BFS}) for branching exploration.
Chain and tree methods share the same Policy and Transition—tree search adds a 
RewardModel for path selection.
This component sharing enables fair comparison: identical domain logic evaluated 
under different inference strategies.

\paragraph{Extending Search Algorithms.}
All tree search algorithms inherit from a \code{BaseTreeSearch} abstract base class 
that handles peripheral concerns—node ID management, root creation, 
checkpoint I/O, runtime limits, terminal node collection, 
and binding search-specific context (e.g., tree depth, iteration index) 
to LLM call records for fine-grained cost analysis—so that 
subclasses implement only the \code{search()} method containing 
pure algorithm logic.
New algorithms are registered via \code{@register\_search("name")}, 
which wraps the class into a callable that \code{main\_search.py} 
invokes through a unified interface.
Custom algorithms (e.g., beam search variants, Monte Carlo variants) 
automatically work with all registered components and task types, 
because \code{lits.structures} standardizes the interface 
between search algorithms and components.


%% file: sections/3.2.lits_tool.tex
\subsection{Tool Use}
LiTS adopts a LangChain-compatible tool protocol: each tool is a \code{BaseTool} subclass declaring \code{name}, \code{description}, \code{args\_schema}, and a \code{\_run()} method.
This ensures heterogeneous actions (SQL queries, API calls, geospatial lookups) share a consistent callable interface, so the same \code{ToolUsePolicy} and \code{ToolUseTransition} invoke them interchangeably.
Tools are instantiated with a backend \code{Client} that encapsulates I/O logic, decoupling reasoning from infrastructure.
At runtime, a list of \code{Tool} instances is passed to the agent; the policy selects tools and the transition executes them, grounding abstract reasoning into concrete world interactions.
For benchmarks whose tools carry per-example mutable state (e.g., a knowledge-graph variable tracker that differs per question), the resource registry supports an optional \code{prepare\_example} callback that resets tool internals before each example---stateful tools are natively supported by \code{lits-chain} and \code{lits-search} without benchmark-specific CLI modifications.


%% file: sections/4.eval.tex
\section{Demonstrations and Results}
\label{sec:demo}
To demonstrate \textsc{LiTS}'s generality and extensibility, we evaluate on three task categories, each showcasing different extension mechanisms.
Our goal is to validate component reusability rather than benchmark performance; see Appendix~\ref{app:exp_setup} for full settings and Appendix Table~\ref{tab:component_abstraction} for component configurations.
Environment-grounded and tool-use experiments use Claude 3.5 Sonnet via AWS Bedrock API, so we report cost; language-grounded experiments use self-deployed Llama3-8B, so we report wall-clock time.

LiTS provides a \textit{CLI-first} workflow where all run artifacts
(configs, logs, checkpoints, evaluation outputs) are written to a single
\code{save\_dir}, enabling post-hoc evaluation via
\code{eval --save\_dir} without re-specifying configuration (Appendix~\ref{app:artifacts}).
Search artifacts are saved incrementally—terminal nodes, paths, and intermediate states—so
evaluation can be reproduced without re-running search.
We report accuracy and efficiency metrics captured by \textsc{LiTS}'s \code{InferenceLogger}.

\subsection{Environment-Grounded Tasks: Adding New Domains}
\label{sec:env_grounded}
BlocksWorld is a domain tested in RAP (MCTS) \citep{hao-etal-2023-reasoning}, re-implemented under LiTS.
To add Crosswords, a domain with different state representations,
users register three components sharing the key \code{"crosswords"}:
a Transition subclass, domain-specific prompts, and a dataset loader.
The same \code{EnvGroundedPolicy} and \code{EnvGroundedPRM} are reused.

\begin{lstlisting}[language=Python, numbers=none]
@register_transition("crosswords", task_type="env_grounded")
class CrosswordsTransition(EnvGroundedTransition):
    @staticmethod
    def goal_check(goals, env_state): ...
    @staticmethod  
    def generate_actions(env_state): ...
    def _step(self, state, action, goals, **kwargs): ...

@register_user_prompt('policy', 'env_grounded', 'crosswords')
def crosswords_policy_prompt():
    return PromptTemplate(template="...", ...)

@register_dataset("crosswords", task_type="env_grounded")
def load_crosswords(data_file=None, **kwargs): ...
\end{lstlisting}
Switching from BlocksWorld to Crosswords requires only
\code{--dataset crosswords} on the CLI—the same agent works automatically.

For search, we use 10 iterations with branching factor 3 for BlocksWorld (max depth 6, early termination on first solution) and 30 iterations for Crosswords (max depth 10, no early termination), with rollout depth matching max search depth in both cases. All these can be set in the \code{lits-search} CLI command.

\begin{table}[h]
\centering
\small
\begin{tabular}{ 
    >{\raggedright\arraybackslash}p{1.4cm}
    >{\raggedleft\arraybackslash}p{0.65cm} 
    >{\raggedleft\arraybackslash}p{0.8cm} 
    >{\raggedleft\arraybackslash}p{0.8cm} 
    >{\raggedleft\arraybackslash}p{1.75cm} 
    }
\toprule
\textbf{Task} & \textbf{Method} & \textbf{Out Tok} & \textbf{Cost} & \textbf{Acc} \\
\midrule
\multirow{2}{*}{BlocksWorld} 
  & Chain & 17K & \$1.48 & 26.7\% \\
  & MCTS & 488K & \$21.99 & 66.7\% \\
\midrule
\multirow{2}{*}{Crosswords} 
  & Chain & 2.5K  & \$0.28 & 6.67\%/10.33\%\textsuperscript{*} \\
  & MCTS & 14K & \$2.42 & 0\%\/22.67\%\textsuperscript{*} \\
\bottomrule
\end{tabular}
\caption{Environment-grounded results (30 examples). Same components across domains. \textsuperscript{*}Crosswords accuracy: exact match (all 10 clues correct) / partial match (average clue accuracy).}
\end{table}

\paragraph{Mode Collapse in Infinite Action Spaces.}
Despite temperature escalation (0.8$\rightarrow$1.2) upon duplicate detection, we observe an 81.1\% duplicate rate across 127 LLM calls on Crosswords (Table~\ref{tab:llm-diversity}), with nearly identical rates among incorrect outputs.
Even with oracle rewards from ground-truth answers, tree search fails---confirming the bottleneck is action diversity, not reward quality.
While our measurements are on a single environment, the cause is structural: in open-ended text action spaces, the policy is an LLM with no calibrated stochastic policy head, and temperature-based sampling explores at the token level rather than the action level. Two LLM calls with the same prompt therefore tend to produce semantically duplicated actions even at high temperature, regardless of domain.
In contrast, finite action spaces (e.g., BlocksWorld) guarantee branching diversity via deterministic fallback to unselected valid actions.

\begin{table}[h]
\centering
\small
\begin{tabular}{lc}
\toprule
\textbf{Metric} & \textbf{Value} \\
\midrule
Unique states visited & 16 \\
Avg. policy calls per state & 7.9 \\
Duplicate rate (all) & 81.1\% \\
Duplicate rate (incorrect) & 81.0\% \\
Correct outputs & 17.3\% \\
\bottomrule
\end{tabular}
\caption{LLM action diversity on Crosswords with temperature escalation ($T$: 0.8$\rightarrow$1.2). Near-identical duplicate rates for all outputs vs.\ incorrect-only outputs indicate temperature scaling fails to improve exploration.}
\label{tab:llm-diversity}
\end{table}

\subsection{Tool-Use Tasks: Resource Registration}
\label{sec:tool_use}

For tool-use tasks, users register two functions sharing the same key:
a dataset loader (via \code{@register\_dataset}) and a resource loader
(via \code{@register\_resource}) that returns the tools and context
the agent can use during search.

\begin{lstlisting}[language=Python, numbers=none]
@register_dataset("mapeval-sql", task_type="tool_use")
def load_mapeval_sql(**kwargs):
    return [{"question": ..., "answer": ...}, ...]

@register_resource("mapeval-sql")
def load_mapeval_sql_resource(**kwargs):
    return {
        "tools": [QuerySQLTool(db), ...],
        "tool_context": "...",
    }
\end{lstlisting}

Tools follow a LangChain-compatible \code{BaseTool} protocol
(\code{name}, \code{description}, \code{args\_schema}, \code{\_run()});
users can use built-in tools or define their own subclasses.
The generic \code{ToolUsePolicy} and \code{ToolUseTransition}
handle orchestration---no component flags are needed on the CLI.

\begin{table}[h]
\centering
\small
\begin{tabular}{
    >{\raggedleft\arraybackslash}p{0.8cm}
    >{\raggedleft\arraybackslash}p{0.8cm}
    >{\raggedleft\arraybackslash}p{0.8cm}
    >{\raggedleft\arraybackslash}p{1.75cm}
    }
\toprule
\textbf{Out Tok} & \textbf{Inv} & \textbf{Cost} & \textbf{Acc} \\
\midrule
10.6K & 62 & \$0.57 & 40\% \\
\bottomrule
\end{tabular}
\caption{MapEval-SQL ReAct results (10 examples).
\textbf{Inv} = total LLM invocations.}
\label{tab:tool_use}
\end{table}

On MapEval-SQL (10 examples), ReAct achieves 40\% (Table~\ref{tab:tool_use}).
We ran MCTS on a 3-example subset following the LATS reward design~\citep{zhou2024language}; at \$18.40 for 3 examples (\$6.13/example vs.\ \$0.05/example for ReAct), it achieved 0\%.
Self-preference bias~\citep{chen-etal-2025-beyond} in the LLM-as-judge reward model leads MCTS to favor verbose but incorrect queries.
This contrasts with \S\ref{sec:env_grounded}, where ground-truth environment rewards enable effective tree search—highlighting that reward model quality is the bottleneck for tool-use tree search.

\subsection{Language-Grounded Tasks: Composability and Extensibility}
\label{sec:lang_grounded}

This subsection demonstrates three levels of extensibility on MATH500.

\paragraph{Component Reuse Across Algorithms.}
ReST-MCTS*~\citep{zhang2024restmcts} uses built-in components—ConcatPolicy,
ConcatTransition, and GenerativePRM—requiring no user code.
Since the original PRM is unreleased, we substitute the publicly
available PRM from \citet{xiong2024rlhflowmath}, illustrating how
decoupled components enable reproduction with missing pieces.

\paragraph{Custom Formulations via Registered Components.}
RAP~\citep{hao-etal-2023-reasoning} requires a different reasoning
structure: sub-question decomposition instead of step-by-step
concatenation.
Users register custom components that redefine the Step structure
and the Policy/Transition logic:

\begin{lstlisting}[language=Python, numbers=none]
@dataclass
class SubQAStep(Step):
    sub_question: str = ""
    sub_answer: str = ""

@register_policy("rap")
class RAPPolicy(Policy):
    """Generates candidate sub-questions."""
    def get_actions(self, state, **kw): ...

@register_transition("rap")
class RAPTransition(Transition):
    """Answers sub-questions, updates state."""
    def _step(self, state, action, **kw): ...
\end{lstlisting}

The MCTS algorithm itself needs no code changes—only parameter tuning.
This follows the same decorator pattern as \S\ref{sec:env_grounded},
but registers Policy, Transition, and RewardModel instead of just
a Transition.

\paragraph{Custom Search Algorithms.}
BFS~\citep{yao2023tree} is implemented as a registered search algorithm via
\code{@register\_search}, inheriting peripheral concerns from
\code{BaseTreeSearch} (\S3.4):

\begin{lstlisting}[language=Python, numbers=none]
@register_search("bfs", config_class=BFSConfig)
class BFSSearch(BaseTreeSearch):
    def search(self, query, query_idx):
        # Pure algorithm logic: depth-bucketed
        # frontier loop with beam pruning
        ...
\end{lstlisting}

BFS automatically works with all registered components—the same
ConcatPolicy, ConcatTransition, and GenerativePRM used by
ReST-MCTS* require no adaptation.

\paragraph{Setup.}
We evaluate on the first 100 MATH500 examples with numerical answers
using Llama3-8B.
RAP requires a completion model (Llama3-8B base);
ReST and ToT use the instruct variant (Llama3-8B-Instruct).
All tree search methods use 10 iterations,
branching factor 3, and temperature 0.7--0.8
(see Appendix~\ref{app:exp_setup} for full settings).

\begin{table}[h]
\centering
\small
\begin{tabular}{lcccc}
\toprule
\textbf{Method} & \textbf{Acc} & \textbf{Out Tok} & \textbf{Inv} & \textbf{Time} \\
\midrule
CoT & 17\% & 12.9K & 100 & 0.6h \\
RAP (MCTS)\textsuperscript{\dag} & 18\% & 4.47M & 3.6K & 8.0h \\
ReST (MCTS) & 37\% & 2.24M & 4.0K & 26.0h \\
ToT (BFS)\textsuperscript{\ddag} & 39\% & 1.53M & 2.8K & 14.7h \\
\bottomrule
\end{tabular}
\caption{MATH500 results (100 examples, Llama3-8B).
\textsuperscript{\dag}User-registered components;
\textsuperscript{\ddag}user-registered search algorithm;
unmarked rows use only built-in components and algorithms.
\textbf{Inv} = total LLM invocations;
\textbf{Time} = total wall-clock time including all LLM calls.}
\label{tab:lang_grounded}
\end{table}

\paragraph{Analysis.}
ReST (MCTS) and ToT (BFS) share identical built-in components,
isolating the effect of the search algorithm:
BFS achieves comparable accuracy (39\% vs.\ 37\%) in roughly
half the wall-clock time, demonstrating that a user-registered
algorithm immediately benefits from existing components.
RAP, despite using MCTS with the same iteration budget,
achieves only 18\%—its sub-question decomposition formulation
is less effective on MATH500 than step-by-step reasoning,
illustrating how component choice dominates algorithm choice.
All three tree search methods improve over CoT (17\%),
confirming that branching exploration helps even with a small model.

%% file: sections/2.related.tex
\section{Related Work}
\label{sec:related}

\begin{table}[t]
\centering
\small
\begin{tabular}{lccc}
\toprule
\textbf{Feature} & \textbf{LiTS} & \textbf{LLM-R} & \textbf{LG} \\
\midrule
Task-agnostic search & \cmark & \cmark & \xmark \\
Component sharing & \cmark & \xmark & \xmark \\
Tool-use tree search & \cmark & \xmark & \xmark \\
Prompt registry & \cmark & \xmark & \xmark \\
Decorator extension & \cmark & \xmark & \xmark \\
Inference logging & \cmark & \xmark & \xmark \\
Checkpointing & \cmark & \xmark & \texttildelow \\
\bottomrule
\end{tabular}
\caption{Feature comparison. LLM-R = LLM Reasoners; LG = LangGraph; \texttildelow\ = partial.}
\label{tab:comparison}
\end{table}

We compare \textsc{LiTS} with \textbf{LLM Reasoners} \citep{hao2024llm}, the only other framework providing complete tree search implementations, and \textbf{LangGraph}~\cite{langgraph}, the most widely adopted LLM orchestration framework, which practitioners may consider as an alternative despite lacking native tree search support. Table~\ref{tab:comparison} summarizes key differences.

LLM Reasoners bundles task-specific logic into monolithic configuration classes, requiring users to re-implement full task logic for each search method.
LiTS factors out reusable components—users implement domain-specific functions once.
LangGraph requires custom state types, expand/score/prune functions, and manual graph wiring for tree search; LiTS provides pre-implemented algorithms where the same components work across tasks with only prompt registration.


%% file: sections/5.conclusion.tex
\section{Conclusion}

We presented \textsc{LiTS}, a modular framework that decomposes LLM tree search into reusable components---Policy, Transition, and RewardModel---shared across search algorithms and task types.
A decorator-based registry enables extension without modifying core code: users register domain-specific components, datasets, tools, or search algorithms, and the CLI automatically resolves them by registered name.
Our demonstrations show that the same built-in components work across MCTS and BFS, that adding a new domain requires only a Transition subclass and dataset loader, and that tool-use benchmarks integrate via resource registration.
We also identify that in infinite action spaces, LLM policy diversity---not reward quality---is the primary bottleneck for effective tree search.

\paragraph{Extensibility in practice.}
The modular design has already supported follow-up research without modifying the core: a Branching-Necessity (BN) evaluator that drives the Chain-in-Tree continuation phase~\citep{li2025chain} integrates as a new component type alongside Policy/Transition/RewardModel; a context-augmentation module pluggable at the Policy layer is currently under active development for cross-trajectory memory and self-reflection. Both extensions reuse the same data structures and search loop without changes.

\section*{Reproducibility Statement}
Every run writes a single \code{config.json} containing all resolved CLI flags, component selections, and per-component arguments (including defaults) to its \code{save\_dir}, alongside seeded sampling and saved checkpoints (\S\ref{sec:demo}, Appendix~\ref{app:artifacts}). Evaluation can then be reproduced from saved terminal nodes alone via \code{lits-eval --save\_dir}, which auto-loads the config from \code{save\_dir/config.json} without re-running search. Because the saved \code{config.json} captures every parameter, any reported run can also be re-executed by passing the same flags back to \code{lits-search}.

\section*{Limitations and Future Work}
\textsc{LiTS}'s scope is intentionally narrow at present, with several directions left for follow-up work:
\begin{itemize}
\item \textbf{Empirical scale.} Our evaluation is demonstration-focused; rigorous benchmarking across larger datasets and a wider range of model sizes is left for downstream studies that build on \textsc{LiTS}.
\item \textbf{Mode-collapse generality.} The finding is based on a single environment (Crosswords). Systematic study across decoding strategies (nucleus sampling variants, temperature schedules, structured generation constraints) and additional open-ended domains is needed to establish generality.
\item \textbf{Tool-use reward calibration.} The negative result reflects a known limitation of LLM-as-judge reward models~\citep{chen-etal-2025-beyond} rather than a framework limitation, but motivates integrating calibrated verifiers and trained PRMs as future work.
\item \textbf{Algorithm coverage.} Only MCTS and BFS ship as built-ins; A* and beam-search variants are natural extensions via \code{@register\_search} and are planned.
\item \textbf{Throughput.} Tree search at scale requires concurrent and batched LLM calls; this is an ongoing engineering direction.
\item \textbf{Tool ecosystem.} The current \code{BaseTool} interface (compatible with LangChain) requires each tool to be implemented as a Python class and registered before search begins via \code{@register\_resource}. Adding a new tool therefore requires editing and reloading the user's Python module. We plan to integrate the Model Context Protocol (MCP), under which tools are exposed by separate MCP servers and an agent connects via an MCP client to enumerate available tools and invoke them through a standard JSON-RPC interface. With this, users can attach a third-party MCP server (e.g., for databases or file systems) directly from the CLI, and tool updates propagate without code changes in \textsc{LiTS}.
\end{itemize}


%% file: sections/6.1.app-save-dir.tex
\begin{table*}[h!]
\centering
\small
\begin{tabular}{p{4cm} p{10cm}}
\toprule
\textbf{Search (\code{lits\_search})} &
\textbf{\code{blocksworld\_rap/run\_0.2.5/}} \\
\midrule
\code{checkpoints/} &
\code{\{query\_idx\}\_\{iter\}.json}: intermediate tree states per iteration;\quad
\code{\{query\_idx\}\_result.json}: final result with full tree path.\\
\code{terminal\_nodes/} &
\code{terminal\_nodes\_\{query\_idx\}.json}: all terminal nodes found.\\
\code{config.json} & Search configuration (many params).\\
\code{execution.log} & Execution log.\\
\code{eval.log} & Evaluation log.\\
\code{inferencelogger.log} & LLM inference log (token/latency).\\
\code{treetojsonl.jsonl} & Selected paths for visualization.\\
\midrule
\textbf{Chaining (\code{lits\_chain})} &
\textbf{\code{blocksworld\_chain/run\_0.2.5/}} \\
\midrule
\code{checkpoints/} &
\code{\{query\_idx\}.json}: single trajectory per query.\\
\code{config.json} & Simple config (fewer params).\\
\code{execution.log} & Execution log.\\
\code{eval.log} & Evaluation log.\\
\code{eval\_results.json} & Aggregated evaluation results.\\
\code{inferencelogger.log} & LLM inference log (token/latency).\\
\bottomrule
\end{tabular}
\caption{Per-run artifact layout for search and chaining in LiTS. Each run writes all outputs under a single \code{save\_dir}, enabling post-hoc inspection and evaluation via \code{eval --save\_dir}.}
\label{tab:artifacts}
\end{table*}

\section{Run Artifacts and Directory Structure}
\label{app:artifacts}
LiTS writes all run artifacts to a single \code{save\_dir}, enabling post-hoc evaluation via \code{eval --save\_dir} without re-specifying configuration.
Table~\ref{tab:artifacts} summarizes the artifact layout for search and chaining runs.

\paragraph{Polymorphic Serialization.}
A central \code{TYPE\_REGISTRY} resolves polymorphism across \code{Action}, \code{Step},
and \code{State} subclasses.
The \code{state.save()} method serializes trajectories into JSON with
\code{\_\_type\_\_} metadata, 
ensuring that heterogeneous traces—such as a ReAct
trajectory containing interleaved thoughts, tool calls, and environment
observations—are preserved with full fidelity. 
\code{state.load()} utilizes the registry to dynamically reconstruct the exact
Python dataclasses from these JSON payloads.

%% file: sections/6.2.app-exp-additional.tex
\input{tabs/abstraction}
\input{tabs/component_configs}
\section{Experimental Setup}
\label{app:exp_setup}
Below are the setups for the three types of tasks: (1) \textit{language-grounded reasoning} using MATH500 (100 examples), (2) \textit{environment-grounded planning} using BlocksWorld and Crosswords (30 examples each), and (3) \textit{tool-use tasks} using MapEval-SQL (10 examples). 

As a demonstration paper, our goal is to validate component reusability across task types rather than benchmark state-of-the-art performance.
The sample sizes are sufficient to demonstrate that the same components work across algorithms (MCTS vs.\ BFS) and domains (BlocksWorld vs.\ Crosswords), which is the core claim of this work.
Tree search methods are computationally expensive—each search iteration requires multiple LLM calls for policy sampling and reward evaluation—so we use representative subsets that balance demonstration coverage with practical cost.

Environment-grounded (\S\ref{sec:env_grounded}) and tool-use (\S\ref{sec:tool_use}) experiments use Claude 3.5 Sonnet via AWS Bedrock.
Language-grounded experiments (\S\ref{sec:lang_grounded}) use self-deployed Llama3-8B: the base model for RAP (which requires a completion model) and Llama3-8B-Instruct for ReST-MCTS* and ToT-BFS.
All tree search methods use $n\_actions=3$; see per-task settings in \S\ref{sec:env_grounded}, \S\ref{sec:tool_use}, and \S\ref{sec:lang_grounded}.

\section{Additional Tables and Interfaces}
\label{app:additional}

\paragraph{BaseTool Interface.}
Each tool extends \code{BaseTool} with three attributes and one method:
\begin{lstlisting}[language=Python, numbers=none]
class QuerySQLTool(BaseTool):
    name = "query_sql"
    description = "Execute SQL query"
    args_schema = QuerySQLInput  # Pydantic model
    def _run(self, query: str) -> str: ...
\end{lstlisting}

%% file: tabs/abstraction.tex
\begin{table*}[htbp]
    \centering
    \resizebox{\textwidth}{!}{%
    \begin{tabular}{@{}lccc@{}}
        \toprule
        \textbf{Task Type} & \textbf{Action} & \textbf{Step} & \textbf{State} \\ 
        \midrule
        \textbf{Language-Grounded} & \texttt{StringAction} & \texttt{ThoughtStep} / \texttt{SubQAStep} & \texttt{TrajectoryState} \\
        \textit{(e.g., Math500, GSM8K)} & \small{Free-form text generation} & \small{Reasoning thought or sub-QA pair} & \small{Sequence of reasoning steps} \\ 
        \addlinespace
        \textbf{Tool-Use} & \texttt{ToolUseAction} & \texttt{ToolUseStep} & \texttt{ToolUseState} \\
        \textit{(e.g., MapEval)} & \small{Structured tool call (JSON)} & \small{Thought, Action, Observation} & \small{Trajectory with tool history} \\ 
        \addlinespace
        \textbf{Environment-Grounded} & \texttt{EnvAction} & \texttt{EnvStep} & \texttt{EnvState} \\
        \textit{(e.g., BlocksWorld)} & \small{Valid environment command} & \small{Action and resulting state} & \small{Trajectory + Env Snapshot} \\ 
        \bottomrule
    \end{tabular}%
    }
    \caption{Structure subclasses across task types. Each task type defines polymorphic Action, Step, and State subclasses that carry task-specific data while exposing a uniform interface to search algorithms.}
    \label{tab:structure_abstraction}
\end{table*}

%% file: tabs/component_configs.tex
\begin{table*}[htbp]
    \centering
    \resizebox{\textwidth}{!}{%
    \begin{tabular}{@{}llcccc@{}}
        \toprule
        \textbf{Task Type} &
        \textbf{Execution} &
        \textbf{Method} &
        \textbf{Policy} &
        \textbf{Transition} &
        \textbf{Reward Model} \\
        \midrule
        
        \multirow{3}{*}{\textbf{Environment-Grounded}} 
        & Chain 
        & EnvChain 
        & EnvGroundedPolicy 
        & BlocksWorldTransition\textsuperscript{\dag} 
        & -- \\
        
        & Tree 
        & MCTS / BFS 
        & EnvGroundedPolicy 
        & BlocksWorldTransition\textsuperscript{\dag} 
        & EnvGroundedPRM \\

        & Tree 
        & MCTS / BFS 
        & EnvGroundedPolicy 
        & CrosswordsTransition\textsuperscript{\dag} 
        & EnvGroundedPRM \\
        
        \addlinespace
        
        \multirow{2}{*}{\textbf{Tool-Use}} 
        & Chain 
        & ReActChat 
        & ToolUsePolicy 
        & ToolUseTransition 
        & -- \\
        
        & Tree 
        & MCTS / BFS 
        & ToolUsePolicy 
        & ToolUseTransition 
        & ToolUsePRM \\
        
        \addlinespace
        
        \multirow{2}{*}{\textbf{Language-Grounded}} 
        & Tree 
        & MCTS 
        & RAPPolicy\textsuperscript{\dag} 
        & RAPTransition\textsuperscript{\dag} 
        & RapPRM\textsuperscript{\dag} \\
        
        & Tree 
        & MCTS / BFS 
        & ConcatPolicy 
        & ConcatTransition 
        & GenerativePRM \\
        
        \bottomrule
    \end{tabular}
    }
    \caption{Component configurations used in our demonstrations. Unmarked components are built-in (shipped with \textsc{LiTS}); \textsuperscript{\dag}components are user-registered via \code{@register\_*} decorators. The same built-in components are reused across methods within a task type; only the RewardModel is added for tree search.}
    \label{tab:component_abstraction}
\end{table*}